# Real-time smart vehicle surveillance system

[1]Shantha Kumar S, [2]Vykunth P, [3]Jayanthi D
Sri Venkateswara College of Engineering,
{[1]2017it0702,[2]2017it0744,[3]jayanthi}@svce.ac.in

*Abstract-* Over the last decade, there has been a spike in criminal activity all around the globe. According to the Indian police department, vehicle theft is one of the least solved offenses, and almost 19% of all recorded cases are related to motor vehicle theft. To overcome these adversaries, we propose a real-time vehicle surveillance system, which detects and tracks the suspect vehicle using the CCTV video feed. The proposed system extracts various attributes of the vehicle such as Make, Model, Color, License plate number, and type of the license plate. Various image processing and deep learning algorithms are employed to meet the objectives of the proposed system. The extracted features can be used as evidence to report violations of law. Although the system uses more parameters, it is still able to make real time predictions with minimal latency and accuracy loss.

## I.INTRODUCTION

In recent years, there is a spike in vehicle related criminal activities. Police data show that motor vehicle theft is one of the least-solved crimes. In 2018, the total number of motor vehicle theft cases reported stood at 44,158, as compared to 39,084 in 2017 – which accounted for about 19% of the total crimes reported. However, only 19.6% cases were solved. The problem with vehicle surveillance is that it requires a lot of manual work. Automating this tedious process could save a lot of time and manual work. This research proposes a real-time surveillance system which makes use of various attributes like Make, Model, Color, License plate number, and type of the license plate to identify the suspect vehicle. Various computer vision and Artificial intelligence algorithms are explored in the process.

*Artificial intelligence*

(AI) refers to the simulation of human intelligence in machines that are programmed to think like humans and mimic their actions. The term may also be applied to any machine that exhibits traits associated with a human mind such as learning and problem-solving. The ideal characteristic of artificial intelligence is its ability to rationalize and take actions that have the best chance of achieving a specific goal. Deep learning techniques enable this automatic learning through the absorption of huge amounts of unstructured data such as text, images, or video. Artificial intelligence is based on the principle that human intelligence can be defined in a way that a machine can easily mimic it and execute tasks, from the simplest to those that are even more complex. The goals of artificial intelligence include learning, reasoning and perception.

The research is structured as follows, section 2 reviews the literature, section 3 briefs about the research gaps in the existing work while section 4 states the motivation behind this research. Section 5 puts forth the objective of the research and section 6 explains about the system design and description of modules. Section 7 discusses the results. Section 8 finally concludes the research.

## II. BACKGROUND

Bochkovskiy et al. [1] proposed the YOLOv4, which improves upon YOLOv3 and offers a state-of-the-art detector that is faster and more accurate than other object detectors. Nafzi M et al. [2] proposed a way to collect and label a large-scale data set for vehicle identification based on make/model and colour classification using Deep Neural Networks. Colour classification is robust and shows very good results. It could be used by the Automated Vehicular Surveillance (AVS) or by the fast analysis of video data. The make/model classification shows very good results on controlled data and good results on video data. Dehghan et al. [3] proposed a system with a deep convolutional neural network that is not only computationally inexpensive but also provides state-of-the-art results for recognizing the make, model, and colour of vehicles. The combination of Sighthounds novel approach to the design and implementation of deep neural networks and a sizable dataset for training allowed high degrees of accuracy to label vehicles in real-time. This method showed significant improvement in several experiments for both classification and verification tasks but was trained and tested using the Stanford Cars dataset, which consists of only 196 classes of cars. H.Wang et al[4] compared various Deep Learning algorithms for vehicles. Five mainstream deep learning object detection algorithms in vehicle detection, namely the faster R-CNN, R-FCN, SSD, RetinaNet, and YOLOv3 were compared on the KITTI data and the obtained results were analysed. Hendry et al. [5] addressed the problem of car license plate detection using You Only Look Once (YOLO)-darknet deep learning framework. YOLO's 7 convolutional layers are used to detect a single class using a sliding window mechanism to identify each digit of Taiwan's car license plates. Kang Q et al. [6] demonstrates the use of thermal infrared cameras in the field of DAS and discusses their performance in night-time vehicle recognition. The study compares the performances of nine CNNs based on the core structure of SqueezeNet for robust vehicle recognition using thermal infrared images. In this work, a dataset of thermal infrared images comprising four vehicle classes was generated: bus, truck, van, and car. Laroca et al. [7] proposed an efficient and 13 layout independent ALPR system based on YOLO by evaluating and optimizing different models with various modifications, aiming at achieving the best speed/accuracy trade-off at each stage. The system uses a unified approach for license plate detection and layout classification to improve the recognition results using post-processing rules. Lee et al [8] proposed a novel deep learning approach for Vehicle Make and Model Recognition (VMMR) using the SqueezeNet architecture. The memory efficient SqueezeNet architectures employed real-time application use because of compressed model sizes without a considerable change in the recognition accuracy. A unique and robust real-time VMMR system was proposed by Manzoor et al. [9] that can handle the challenges such as image acquisition, variations in illuminations and weather, occlusions, shadows, reflections, a large variety of vehicles, etc using Random Forest (RF) and Support Vector Machine (SVM). The proposed VMMR system recognizes vehicles on the basis of make, model, and generation (manufacturing years) while the existing VMMR systems can only identify the make and model. N. Dhieb et al [10] proposed an automated and efficient solution for vehicle damage detection and localization using deep learning. The proposed solution combines deep learning, instance segmentation, and transfer learning techniques for features extraction and damage identification. Its objective is to automatically detect damages in vehicles, locate them, classify their severity levels, and visualize them by contouring their exact locations.

### III. RESEARCH GAP

The existing systems use only one or two attributes of the vehicle to recognize it. The Vehicle detection system uses SSD and YOLOv3 which is not that much suitable for real-time processing, since there exist better models like YOLOv4. The current state-of-the art vehicle recognition model uses SqueezeNet with an inference time of 0.52 ms and accuracy of 97% but it requires infrared cameras Make model recognition uses Stanford dataset which contains only 196 classes. The VMMR system is based on SVM-VMMR with a processing speed 10.1 FPS For the existing VMMR, they have good accuracy for Colour classification, but the model is not able to perform very well due to the large number of classes. License plate detection system uses post-processing rules to improve results and achieved an average recognition rate of 98% and up to 30 FPS for detection only. The sliding-window mechanism achieved a plate detection accuracy of 98.22% and 78% for plate recognition with an inference time of 800ms to 1s for each input image. The current system for Vehicle Damage detection and localization uses Inception Resnet V2 pre-trained model and achieves very high accuracies but is not suitable for real-time processing.

### IV. MOTIVATION

The crime rate in India has increased significantly in the last decade. According to statistics the crime rate grows by 1.6% every year. The number of crimes committed in 2019 alone is around 50,74,635 out of which the crime related to car theft is 104,401. This is almost 13 cases in 100,000 cases. This increase in the number of crimes and thefts demands more authorities to be involved. This need for more authorities not only takes a huge toll on spending the government funds but also is no match to the exponentially increasing car related crimes. This alarming situation opens the gate for innovation through technological advancements. The increasing technology in the field of artificial intelligence and machine learning allows us to automate the process of detecting crime and its related activities. All the existing models use only few attributes of the vehicle and does not provide a complete solution for automated surveillance. Also, an efficient system is needed that combines all these models and at the same time achieve real-time processing speeds.

### V. OBJECTIVE

The proposed system uses multiple models for a complete vehicle surveillance. Even if one or two attributes are not clear, the system can still recognize the vehicle. This work focuses on recognising the stolen vehicle by using the CCTV footage captured. An automated vehicle surveillance system is proposed, using Artificial Intelligence, to recognize and track vehicles by extracting various attributes of the vehicle such as Make, Model, Colour, License plate number, and type of the plate. Various Computer Vision and Deep Learning algorithms will be explored throughout this work. A real-time database is compiled with all these details along with the date, time, and location stamp of the vehicles which can be used to uniquely identify the suspect vehicles involved in criminal activity. Also, the extracted features can be used as evidence to report violations of law such as illegal number plates or expired documents. Even if one or more of the features of the vehicle are not clearly visible, the system can still recognize the vehicle using other features. Based on the recently detected locations, the route of the suspect can be tracked, and necessary legal actions can be taken to prevent the future crimes. The project is divided into modules as listed below.

Fig. 1 depicts the architecture diagram of the proposed Vehicle Surveillance System.

A. *Vehicle Make and Model Recognition*
B. *Vehicle Color Detection*
C. *License plate Recognition*

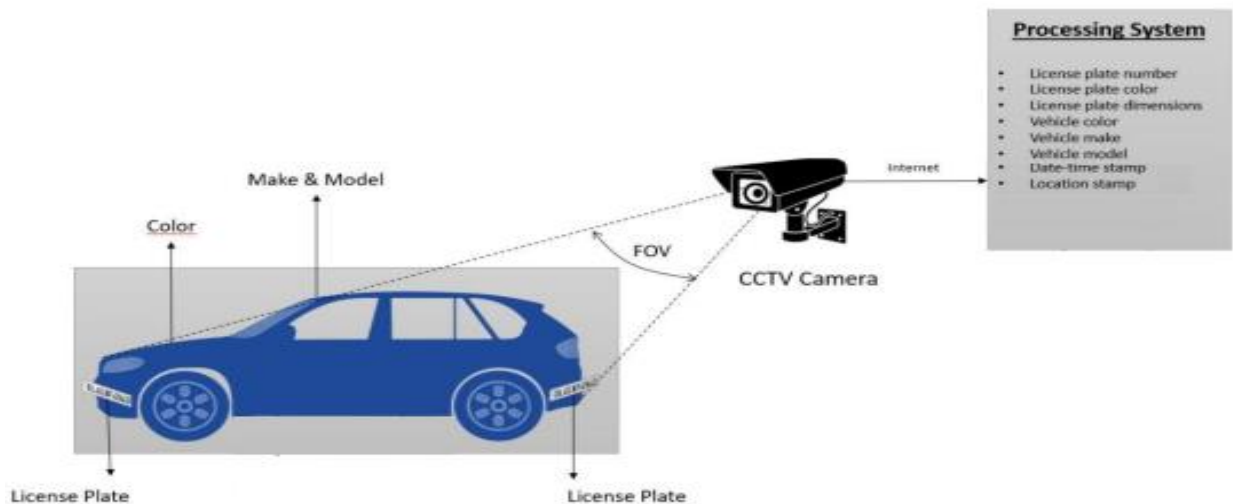

**Fig.1 Vehicle Surveillance System Architecture**

## VI. METHODOLOGY

Surveillance cameras are being used in many applications for monitoring human behaviour. In streets and public roads, cameras are on tolls or for traffic monitoring, for example. These are scenarios on which they extract images that contain the raw data about the vehicles that are circulating. Through the usage of image processing techniques, characteristics of the vehicles like make and model, colour, license plate can be extracted. This information can be used for many purposes in intelligent transportation systems. For example, from the visual identity of the vehicle, obtained by means of the combination of the extracted features, one can verify the vehicle against an expected identity and check for vehicle cloning or appearance changes. Also, by automatically detecting vehicle make and model, it is possible to index them in parking systems.

A. *Vehicle Make and Model Recognition*

This section elucidates about the detection of vehicle make and model. This module makes use of CompCars dataset and MobileNet algorithm to achieve the desired results.

*Collecting Dataset*

The CompCars dataset contains two types of vehicle images: web-nature that are images collected from forums, public websites, and search engines; and surveillance-nature images collected from surveillance cameras. In particular, the web-nature set contains 163 car makes with 1716 car models, with a total of 136726 images. These pictures contain the entire vehicles in various viewpoints and inserted into different backgrounds. For the fine-grained classification task, these makes, and models were combined into a subset of 431 make-model classes, with vehicles of the same model type but produced in different years assigned to the same class. This subset contains 52083 images and is divided into 70% (36456 images) for training, and

30% (15627) for testing. Examples of images from this dataset can be viewed in Fig. 2.

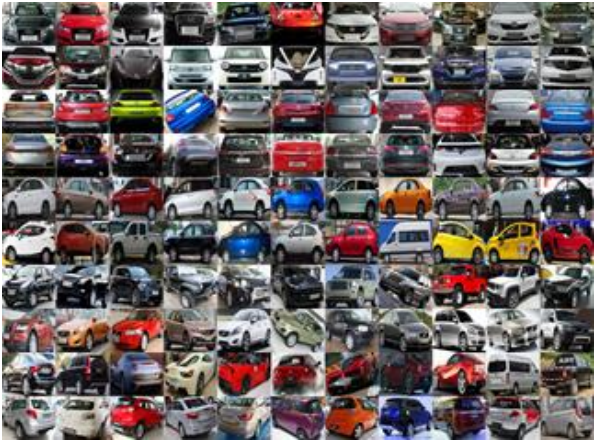

Fig. 2 CompCars Dataset

MobileNets are a class of convolutional neural networks that, by means of depth-wise separable convolutions, achieve lower computational cost during inference. Depth-wise separable convolutions are a form of factorized convolutions; they separate the filtering and the combining operations of the standard convolutions. First, the depth wise convolution applies a single filter to each input channel. These models already have a reduced number of parameters, but hoping to have even faster models, width multiplier and resolution multiplier, can be further tuned so that the user can choose networks with even smaller sizes, though at the cost of reduced accuracy. They have open-sourced 16 models pre-trained on ImageNet, obtained by sweeping these two parameters. The network trained on ImageNet with the highest accuracy uses resolution of 224 by 224 pixels and width multiplier of 1.0 and has approximately 569x106 multiply-add operations.

For each pre-trained model, the last layer was replaced to reflect the number of expected classes. The MobileNet architecture (Fig. 3) has a width multiplier of 1.0 and input size of 224 by 224. To augment the dataset and mitigate the model overfitting on the data the images used during training are first resized to 256 by 256 pixels and then randomly cropped to 224 by 224 pixels. A random horizontal flip was also used for additional data augmentation. Finally, the generator is responsible for delivering the number of samples expected to fit the memory of the used hardware. The batch size contains 32 images that are sampled cyclically from the training set.

The training was made during 71 epochs. The learning rate schedule starts at 0.002 and after the middle of the training it is dropped to 0.0002. Stochastic Gradient Descent was used with a momentum of 0.9, which is the same optimizer used for training GoogLeNet cars.

| Type/Stride | Filter Shape | Input Size |
|---|---|---|
| conv/s2 | 3 x 3 x 3 x 32 | 224 x 224 x 3 |
| Conv dw/s1 | 3 x 3 x 32 dw | 112 x 112 x 32 |
| Conv/s1 | 1 x 1 x 32 x 64 | 112 x 112 x 32 |
| Conv dw/s2 | 3 x 3 x 64 dw | 112 x 112 x 64 |
| Conv/s1 | 1 x 1 x 64 x 128 | 56 x 56 x 64 |
| Conv dw/s1 | 3 x 3 x 128 dw | 56 x 56 x 128 |
| Conv/s1 | 1 x 1 x 128 x 128 | 56 x 56 x 128 |
| conv dw/s2 | 3 x 3 x 128 dw | 56 x 56 x 128 |
| Conv/s1 | 1 x 1 x 128 x 256 | 56 x 56 x 128 |
| Conv dw / s1 | 3 x 3 x 256 dw | 28 x 28 x 256 |
| Conv / s1 | 1 x 1 x 256 x 256 | 28 x 28 x 256 |
| Conv dw / s2 | 3 x 3 x 256 dw | 28 x 28 x 256 |
| Conv / s1 | 1 x 1 x 256 x 512 | 14 x 14 x 256 |
| 5x Conv dw / s1 | 3 x 3 x 512 dw | 14 x 14 x 512 |
| Conv / s1 | 1 x 1 x 512 x 512 | 14 x 14 x 512 |
| Conv dw / s2 | 3 x 3 x 512 dw | 14 x 14 x 512 |
| Conv / s1 | 1 x 1 x 512 x 1024 | 7 x 7 x 512 |
| Conv dw / s2 | 3 x 3 x 1024 dw | 7 x 7 x 1024 |
| Conv / s1 | 1 x 1 x 1024 x 1024 | 7 x 7 x 1024 |
| GlobalAveragePooling2D | Pool 7 x 7 | 7 x 7 x 1024 |
| FC / s1 | 1024 x 431 | 1 x 1 x 1024 |
| Softmax | Classifier | 1 x 1 x 431 |

Fig. 3 MobileNet Architecture for Vehicle Make Model Recognition

### B. Vehicle Color Detection

*Color classification*

K-Means Algorithm K-means is a clustering algorithm used in Machine Learning where a set of data points are to be categorized to 'k' groups. It works on simple distance calculation (1).

$$J(V) = \sum_{i=1}^{c} \sum_{j=1}^{c_i} (\| x_i - v_j \|)^2$$

(1)

where, '$\|x_i - v_j\|$' is the Euclidean distance between $x_i$ and $v_j$.'
'$c_i$' is the number of data points in the cluster. 'c' is the number of cluster centres.

As an image is made of three channels: Red, Green and Blue, each pixel can be represented as a point (x=Red, y=Green, z=Blue) in 3D space and k-means clustering algorithm can be applied on the same. After processing each pixel with the algorithm cluster centroids would be the required dominant colours. For example, the image shown in Fig. 4 can be processed using K-Means algorithm to identify its dominant colours as a 3D plot.

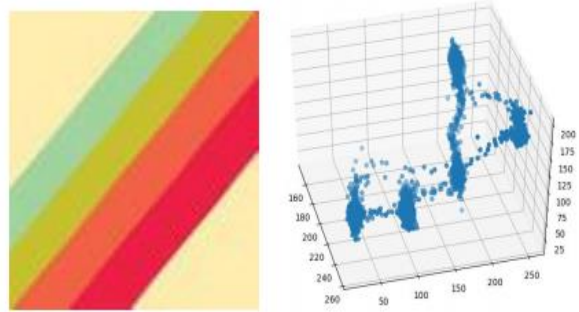

Fig. 4 Plotting the colours in the image

From the plot, it can be easily seen that the data points are forming groups - some places in a graph are denser, which can be considered as different colours' dominance on the image. Such clusters can be achieved through k-means clustering. The figure 5 shows the clusters in the plot.

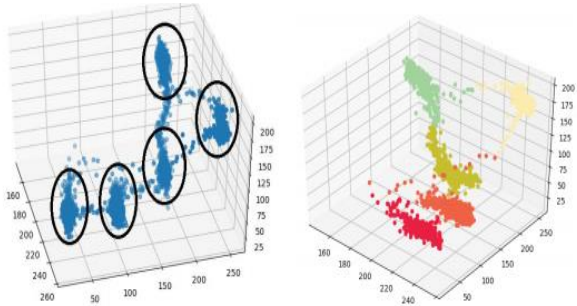

**Fig. 5 Cluster of colours in the plot**

One more add-on, the clusters of colours can be sorted in the order of their dominance, i.e., the clusters with most data points followed by the lesser ones by using histogram. The sorted array of dominant colours is shown in Fig. 6.

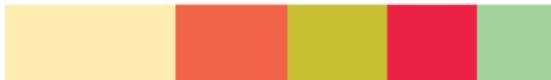

**Fig. 6 Dominant colours in order**

*Vehicle Colour Classification*

The K-Means colour classification algorithm can be applied to the cropper vehicle image from the previous module in order to identify the dominant colour of the vehicle [2]. This is represented in Fig. 6

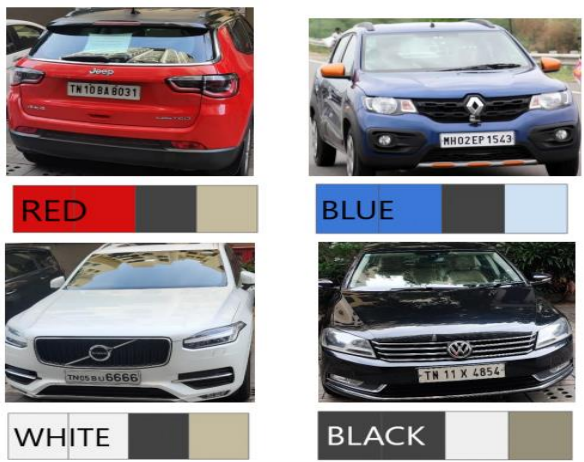

**Fig. 7 Vehicle Color Classification using K-Means clustering**

*C. License Plate Recognition*

The first step is to detect the License plate from the car. The contours feature in OpenCV is used to detect rectangular objects to find the number plate. The accuracy can be improved if the exact size, colour, and approximate location of the number plate is known. Normally the detection algorithm is trained based on the position of the camera and type of number plate used in that particular country. This gets trickier if the image does not even have a car, in this case, an additional step is required to detect the car and then the license plate [12x].

Once the counters have been detected they are sorted from big to small and consider only the first 10 results ignoring the others. The counter could be anything that has a closed surface but of all the obtained results the license plate number will also be there since it is also a closed surface. To filter the license plate image among the obtained results, loop through all the results and check which has a rectangle shape contour with four sides and a closed figure. Since a license plate would definitely be a rectangle four-sided figure. Then Tesseract OCR will be applied. The Tesseract 4.00 includes a new neural network subsystem configured as a text line recognizer. The segmented characters are passed into the tesseract model to predict the characters of the license plate. The working of the license plate recognition module is depicted in Fig.8.

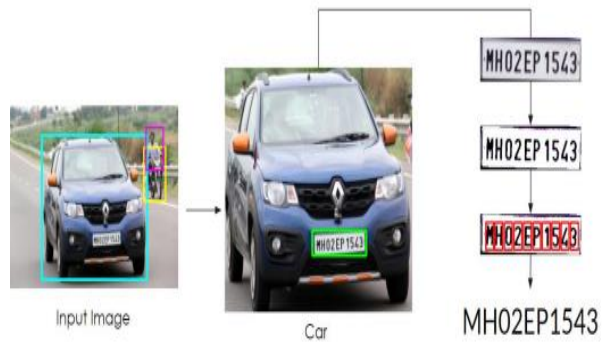

**Fig. 8 Working of License Plate Recognition**

### VII. RESULTS AND DISCUSSION

To analyse the performance of Vehicle Make and Model Recognition (VMMR) module, several tests were performed to examine the results and inspect the influence of the different parameters. The metric used to compare the results was the accuracy (the number of right predictions divided by the total amount of samples). In top 1 if the prediction with the highest probability was correct and in top 5 if the correct prediction is listed on the 5 highest probabilities. First, the input resolution of the MobileNets models was set to 224 by 244 and the width multiplier was set to 1.00.

The results are shown in Table 2 and MobileNet 1.00 224 performed the best among the other models with a Top 1 accuracy of 93.7% and Top 5 accuracy of 98.8%.

**Table 1. Experimentation results of MobileNet VMMR**

| Model | Top 1 Accuracy | Top 5 Accuracy |
|---|---|---|
| GoogleNet cars | 91.2 % | 98.1 % |
| MobileNet 1.00 224 | **93.7 %** | **98.8%** |
| MobileNet 1.00 192 | 92.5 % | 98.5 % |
| MobileNet 1.00 160 | 83.9 % | 95.6 % |
| MobileNet 1.00 128 | 54.2 % | 76.7 % |

Table 2 depicts the performance of the individual modules. It can be noted that the modules achieve a very low inference time. The average latency of the pipeline is 0.167s. This low inference time facilitates the real-time application of the system.

**Table 2. Performance metrics of proposed modules**

| Module | Accuracy | Precision | Recall | F1-score | Specificity | Average Time (s) |
|---|---|---|---|---|---|---|
| Vehicle Detection | 0.95225 | 0.962187 | 0.9415 | 0.951731 | 0.963 | 0.062 |
| License Plate Detection | 0.843 | 0.866453 | 0.811 | 0.83781 | 0.875 | 0.048 |
| Optical Character Recognition | 0.958654 | 0.951923 | 0.964912 | 0.958374 | 0.952562 | 0.034 |
| Colour Classification | 0.866 | 0.932 | 0.823322 | 0.874296 | 0.921659 | 0.023 |
| Overall | 0.904976 | 0.928141 | 0.885183 | 0.905553 | 0.928055 | 0.167 |

## VIII. CONCLUSION AND FUTURE WORK

Automated Vehicle Surveillance is a complex process and requires a lot of factors to be taken into account. Important attributes of the vehicle such as License plate number, Vehicle colour, License plate type have been successfully implemented. The system developed so far works with both images as well as videos and achieved great accuracy of 90.4%. The use of motion detection and tiny architecture of YOLOv4 model along with optimal image processing techniques helped in reducing the inference time to 0.167 seconds per frame, thus making it suitable for real-time surveillance.

As of future work, the main goal is to make the model edge deployable. The processing can be done using cloud servers and powerful GPUs which makes the inference much faster. In this way, the data from multiple CCTV cameras can be processed simultaneously by shifting the processing load to distributed cloud servers. Currently, the system is only able to detect and extract license plate details from the standard car plates. Due to the difference in dimensions and aspect ratio of bike license plates, the system is not able to recognise those. This can be done by training YOLO v4 algorithm on bike license plate dataset. The system can be further improved to make predictions in difficult or challenging scenarios such as low-quality cameras, low light or night-time, rainy, foggy weather conditions. The recent deep learning based super resolution algorithms can be used to improve the clarity of pixels to identify and extract details accurately. Also, the system can be trained to identify damaged license plates and an algorithm can be developed to predict the license plate number from damaged plates. Face recognition algorithms can be integrated into the system to recognize the facial features of the driver (suspect) during crime scenes or traffic violations.

## AUTHORS PROFILE

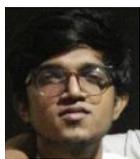
**Shantha Kumar S** - Shantha Kumar has received a bachelor's degree in Information technology from Sri venkateswara college of Engineering in the year 2021. He is a full-time research assistant and is interested in Artificial Intelligence, Machine Learning and Computer vision.

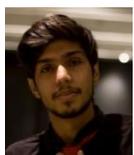
**Vykunth P** - Vykunth has received a bachelor's degree in Information technology from Sri venkateswara college of Engineering in the year 2021. He is a full-time research assistant and is interested in Artificial Intelligence, Machine Learning and Data Science.

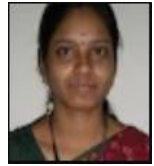
**Jayanthi D**- JAYANTHI D received B.Tech. degree in Information Technology from the Abdul Hakeem College of Engineering, Vellore and M.E degree in Computer Science and Engineering from the Sri Venkateswara College of Engineering, India. She is currently pursuing Ph.D. in Information and Communication at Anna University, Chennai and submitted her thesis. She is currently working as an Assistant Professor at Sri Venkateswara College of Engineering, Sriperumbudur. Her domain of interest includes Cloud Computing, Service Oriented Architecture, Machine learning and Deep Learning. She has published papers in National and International conferences, National and International Journals and SCI Indexed Journals with high impact factor.